%% file: main.tex
\documentclass{article}

\usepackage[final]{neurips_2019}

\usepackage[utf8]{inputenc} 
\usepackage[T1]{fontenc}    
\usepackage{hyperref}       
\usepackage{url}            
\usepackage{booktabs}       
\usepackage{amsfonts}       
\usepackage{nicefrac}       
\usepackage{microtype}      
\usepackage{graphicx}
\usepackage{multirow} 
\usepackage[noblocks]{authblk}
\usepackage{tabulary}

\newcolumntype{K}[1]{>{\centering\arraybackslash}p{#1}}
\title{Finding Social Media Trolls: \\ Dynamic Keyword Selection Methods for Rapidly-Evolving Online Debates}

\author[1]{Anqi Liu}
\author[1]{Maya Srikanth}
\author[2]{Nicholas Adams-Cohen}
\author[3]{R. Michael Alvarez}
\author[1]{Anima Anandkumar}
\affil[1]{Department of Computing and Mathematical Sciences\\
California Institute of Technology}
\affil[2]{Immigration Policy Lab\\Stanford University}
\affil[3]{Division of Humanities and Social Sciences\\California Institute of Technology}
\affil[ ]{\textit{ \{anqiliu@caltech.edu,msrikant@caltech.edu,nadamsco@stanford.edu,rma@caltech.edu, anima@caltech.edu\} }}

\begin{document}

\maketitle

\begin{abstract}
Online harassment is a significant social problem.  Prevention of online harassment requires rapid detection of harassing, offensive, and negative social media posts.  In this paper, we propose the use of word embedding models to identify offensive and harassing social media messages in two aspects: detecting fast-changing topics for more effective data collection and representing word semantics in different domains. We demonstrate with preliminary results that using the GloVe (Global Vectors for Word Representation) model facilitates the discovery of new and relevant keywords  to use for data collection and trolling detection.  Our paper concludes with a discussion of a research agenda to further develop and test word embedding models for identification of social media harassment and trolling.  
\end{abstract}

\input{intro.tex}
\input{relatedwork.tex}

\input{methods.tex}

\input{nextstep.tex}

\subsubsection*{Acknowledgments}

Alvarez thanks the John Randolph Haynes and Dora Haynes for supporting his research in this area.  Prof. Anandkumar is supported by Bren endowed Chair, faculty awards from Microsoft, Google, and Adobe, DARPA PAI and LwLL grants. Anqi Liu is a PIMCO postdoctoral fellow at Caltech.

\subsection*{Codes Repository}
The codes used for generating results in this paper can be accessed from the link: https://github.com/mayasrikanth/TwitterStudiesCode.


\bibliography{bibliography}
\bibliographystyle{apsr}

\end{document}

%% file: intro.tex
\section{Introduction and Background}

Social media data, in particular data from Twitter, are being used in many studies of social, political, and economic behavior \citep{klasnja_etal_2018}. Much of this data is collected by keyword searches, for example, gathering tweets that containt specific keywords or hashtags (e.g., \citet{barbara_etal_2015}).  Methods for collecting social media data using specific and static keywords or hashtags from Twitter are readily available \citep{steinert-threlkeld_2018}, and many researchers now collect and analyze data using specific and static keywords or hashtags.  

This method for collecting social media data is well-suited for monitoring and studying online debate that uses a finite set of researcher-identifiable hashtags or keywords. These situations include conversations with short-lived identifiable hashtags and keywords or where the use of specific terms and phrases is stable in the long term.  In these cases, a researcher can collect either the entire universe of a social media debate, or representative samples of it, for subsequent research.

However, this approach is ill-suited for online debates that are dynamic, where hashtags and keywords in the conversation change rapidly over time.  The terminology that social media communities use to debate particular issues are often highly variable over extended time frames, and communities may use different hashtags and critical phrases as events or individuals change how the issues are discussed. This may be especially true when it comes to monitoring and studying negativity, hate speech, and other forms of abusive social media behavior like ``trolling'' and online harassment.  These types of online conversation may be highly dynamic, as those engaged in these forms of discourse and behavior may parody or mimic otherwise legitimate online conversations, or may alter keyword or hashtag terminology to avoid detection and action by social media platforms themselves.  

In our paper, we explore a new approach for a dynamic keyword search and use it in a preliminary analysis of data from the \#MeToo movement, a social media movement against sexual harassment. Due to the rapidly changing nature of the \#MeToo movement~\citep{manikonda2018twitter}, discovering trolling messages is difficult, as misogynist messages will reflect and respond to changing topics.  Our proposed process involves using a set of seed keywords, pulling a sample of the conversations about the movement, and discovering the words and phrases most associated with this sample to better track the social movement. Specifically, we pull a sample of Twitter data containing the `MeToo' hashtag to get a general sense of conversations about the movement. We also pull top Reddit posts from the Red Pill subreddit (an anti-feminist men's rights activist group) to gather a set of misogynistic speech.

In the next section of our paper, we discuss past research on keyword search before turning to our approach and our preliminary results. We find that leveraging word vectors for keyword detection has the potential to uncover interesting and relevant keywords that may expose previously unseen online interactions. Finally, we conclude with a discussion of next steps in the development and use of our dynamic keyword search methodology. 

\vspace{-0.1cm}

%% file: relatedwork.tex
\section{Related Work}
\vspace{-0.1cm}
There are three approaches used in past work to search social media sites for research-relevant material.  One is to use static keywords:  the researcher defines a set of keywords {\it a priori} using their knowledge of the topic.  This method is easy-to-implement and straightforward and is quite transparent, but it can easily lead to biased data \citep{king_lam_roberts_2017}. If attempting to follow conversations over time, a researcher would need to update this list of static keywords themselves in an entirely non-automated manner. At the other end of the spectrum are automated keyword selection methods that can dynamically update the search keywords, for example, as used by email spam filters.  These are often based on non-transparent machine or deep learning methods, which are difficult for researchers to evaluate.  Moreover, fully-automated keywords extraction methods proposed by the data mining community mainly focus on reducing human labeling effort and improving general-purpose accuracy and coverage for a relatively static topic.~\citep{wang2016identifying,zheng2017semi,wang2016identifying}. However, modeling and keeping track of online debates involving dynamic topics requires detection of language and wording evolution, as well as fast adaptation to new data domains for the learning algorithms. In the middle are semi-automated keywords selection approaches, which involve a combination of human- and machine-based analysis to determine appropriate keywords \citep{king_lam_roberts_2017}; these approaches have the strengths and weaknesses of both fully- and non-automated keyword selection methods.

Our goal is to develop a dynamic keyword searching and updating methodology that is fast and efficient (like fully-automated methods), but which provides transparency and is less biased than non-automated methods.  In a rapidly evolving social media discussion, we believe non-automated keyword selection is inefficient, costly, and time-consuming \citep{golbeck2017}. Furthermore, while semi-automated approaches can improve keyword selection with human intervention, in social media discussions involving negative and offensive language, human involvement in keyword selection has ethical considerations that should be avoided (for example, using human coders to label offensive social media posts \citep{metz2019}). Clustering algorithms instead allow researchers to create large dictionaries of hate and misogynistic speech to locate offensive messages \citep{siegal2019}.

Developing less biased, faster, and more efficient ways to identify social media negativity, hate speech, and online harassment is a significant problem.  According to 2017 survey data from the Pew Research Center, 41\% of Americans have themselves experienced online harassment, and 66\% say that they have seen online harassment directed at others \citep{pew2017}. Of these Americans who have experienced online harassment themselves, 18\% say they have been severely harassed (for example, being stalked, physically threatened, sexually harassed, or harassed over a long period) \citep{pew2017}.

%% file: methods.tex
\section{Our Approach and Preliminary Results}
\vspace{-0.2cm}
We propose to use word embedding models to learn a representation of words from data collected with a static set of keywords. We offer two avenues of keyword extraction: (1) Extract words that co-occur most frequently with the previous set of keywords (2) Use a clustering algorithm to expose "clusters" or "major topics", and for each of these clusters, select a representative keyword using cosine similarity or a ranking algorithm. After collecting data using these keywords as queries and retraining our model, we obtain the next set of queries in a similar fashion. We demonstrate with preliminary results that incorporating word embedding models helps track a dynamic topic evolution. This work paves the way for feature engineering in future trolling detection. Further, by comparing word embeddings from different domains, we discover interesting patterns in trending discussions.

\subsection{The GloVe Word Representation Model}
The GloVe (Global Vectors for Word Representation)~\citep{pennington2014glove} is a renown and commonly used word embedding model. It has gained considerable attention due to its ability to represent linear substructures in data. The GloVe is a log-bilinear model with a weighted least-squares objective, and aims to learn word vectors such that their dot product equals the logarithm of the words' probability of co-occurrence.  In the resulting word vector space, cosine similarity indicates linguistic or semantic similarity between two words, while vector differences capture analogies between pairs of words.  We used the GloVe model to train 50-dimension word vectors on various corpora, which consist of Twitter data obtained using Twitter's standard search API (giving access to tweets posted within the previous 7 days), Wikipedia data, and Reddit data.  The Twitter data used in this analysis was collected over August of 2019. Twitter queries were performed using keywords relevant to the \#MeToo movement. 

{
\begin{figure}[b] 
\begin{center}
\setlength{\tabcolsep}{0pt}
\begin{tabular}{cc}
(a)  & (b)  \\
\includegraphics[trim= {50 20 10 40}, clip, height=5cm]{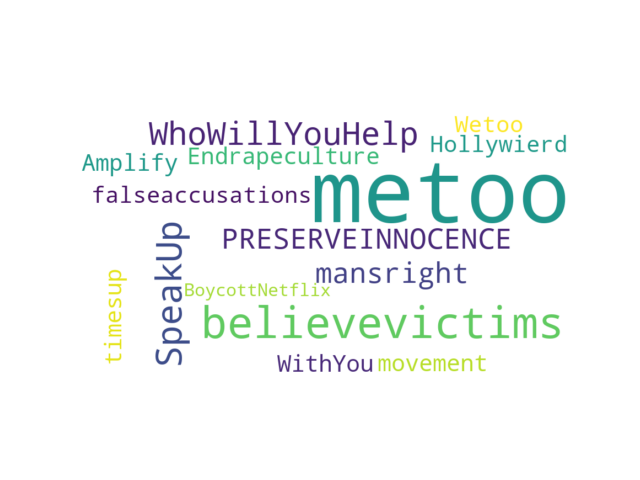} &
\includegraphics[trim= {57 37 10 30}, clip, height=5cm]{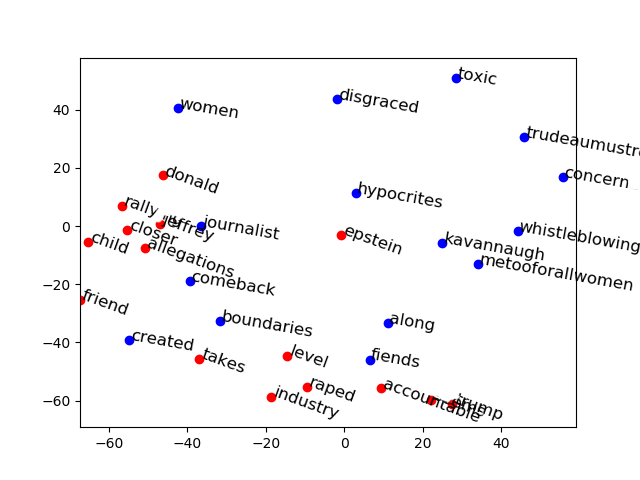}\\
\end{tabular}
\caption{(a)Word cloud containing words with the closest cosine distances to ``MeToo", obtained from a word embedding trained on a subset of \#MeToo data. Larger font corresponds with more proximity to "MeToo"; (b) 2-D visualization of two clusters after conducting K-means clustering on the \#MeToo data with number of clusters set to be 100. Word vectors are reduced to 2-dimension using t-Distributed Stochastic Neighbor Embedding (t-SNE). Blue dots are ``Epstein" cluster while red dots are ``Kavanaugh" cluster.
}
\label{fig:closestwords}
\end{center}
\end{figure}
}

\subsection{Keyword Extraction }
  We visualize the words that are closest linguistically and semantically to a chosen word using the metric of cosine similarity in Figure \ref{fig:closestwords} (a). We can see that the cosine similarity metric can uncover relevant and potentially fruitful keywords to use as queries in the next round of data collection.  

An alternative method for keyword extraction is the application of the K-means clustering~\cite{lloyd1982least} algorithm to the word vector space. Doing so uncovers clusters of words, which are comparable to common discussion topics within a larger topic such as \#MeToo. For example, a news-oriented cluster obtained from \#MeToo data is: 
['Weinstein', 'DonLemon', 'Burke', 'trial', 'Harvey', 'Manhattan', 'Court', 'Justice', 'SexualAssault', 'James', 'fascinating', 'NewYork', 'Supreme', 'SexualPredator', 'wie' (or Women in Engineering), 'Debate', 'untouchable'], while another cluster consists of closely-related hashtags with \#MeToo: ['NotSilent','BelieveWomen', 'WhyIDidntReport', 'EndAbuse', 'TimesUp', 'WomensRights', 'ImWithHer', 'SupportSurvivors', 'Metooindia', 'EnoughIsEnough', 'AcademiaToo', 'NeverAgain', 'WomensMarch', 'MeTooSTEM']. We can then use both cosine similarity and clustering for extracting new keywords. In the future, we expect to develop principled ranking algorithms to establish the relative importance of each candidate keyword, so that we can keep a dynamic ranking of words and avoid exploding of the set.

\subsection{Semantic Representation }
We show two clusters of words discovered by the K-means clustering algorithm in Figure \ref{fig:closestwords} (b). One includes ``Epstein" while the other includes ``Kavanaugh". Both topics are heavily discussed on Twitter and are related with \#MeToo. 

We observe noticeable differences across distinct discussion domains when leveraging word embeddings for keyword extraction. Table \ref{tab:3.3} shows the closest words to ``female" and ``male" for each embedding. Reddit's Red Pill community is a men's rights group and an adversary of the \#MeToo movement. Word embeddings trained on the top 1000 submissions in the Red Pill subreddit as of March 2019 present interesting conceptual differences from word embeddings trained on Twitter data containing the \#MeToo hashtag. We also include embeddings pre-trained on Wikipedia 2014 and Gigaword5 corpus~\citep{pennington2014glove} , which is fairly ``neutral", as a comparison. In the future, making use of embeddings trained on discourse in communities like Red Pill may shed light on trolling and hate speech in other domains.
\begin{table}[h!]
\begin{center}
\begin{tabular}{ |c|lll|lll| } 
\hline
Domains & ``Female" && & ``Male" &&\\
\hline
\multirow{2}{*}{Wikipedia} & adult & young &woman   & female&adult &woman    \\ 
&teenage &girl& individual&girls & individual&older \\ 
+Gigaword5&age&child&older& young &age &child \\
\hline
\multirow{3}{*}{\#MeToo} & companies  & founded  & startups & venture  &Leader & female  \\ 

& desire &  oppressor & employee  & committed & dominated  & junior \\ 

& victims  &  capitalist &  & referred  & day & \\ 
\hline 
\multirow{3}{*}{Red Pill} & sexual  &  negative  & sexuality & alpha & female & plight \\ 

& self & Ability &physical   & attraction & equivalent  & emotional \\ 

& intercourse & respective & dialogue  & beta &  value & sexual \\ 

\hline
\end{tabular}
\end{center}
\caption{The top most similar words to "female" and "male"across different domains.}
\label{tab:3.3}
\end{table}
\vspace{-0.8em}

%% file: nextstep.tex
\section{Discussion and Next Steps}

The method we have discussed here, and which we have pilot tested, shows considerable promise.  Next steps involve further development, better automation of this methodology, and use of a more powerful Twitter API which allow access to a more extensive database of tweets. Then, we plan to test and validate our methodology in several different ways.  One approach is to
compare our methodology to human analysts, to estimate the relative speed and accuracy of our method with respect to human coders.  We also will compare the performance to similar dynamic keyword search methods, for example, \cite{king_lam_roberts_2017}.  Again, critical metrics in that comparison will be how our method compares with respect to both speed and accuracy.  

This study serves as a foundation for a more sophisticated study aiming to better understand the diffusion of information and sentiment through social networks. In our future work, we seek to profile the movement of both information and sentiment through a social network, with the ultimate goal of proposing a mechanism design solution to mitigate abuse and hate-speech targeting individuals and communities online.  We plan to extract network structure that contains an accurate representation of connections between users and their followers with data we can collect. The nodes in the network would be profiled with historical sentiment, predicted with classifiers trained on data we collect using dynamic keyword extraction.  Using both sentiments and network structures, we can characterize the individual and community tendencies within the network. For instance, a node or a community may be classified as an attacker, defender, propagator, or bystander in the discussions in \#MeToo movement. We will identify attributes of the network which facilitate productive and supportive conversations, as well as characteristics that may facilitate abusive interactions. We plan to offer possible methods to encourage empowering discussion while limiting abusive encounters on a social media platform.